\documentclass{article} 
\usepackage{iclr2026_conference,times}

\usepackage{amsmath,amsfonts,bm}

\def\eqref#1{equation~\ref{#1}}

\def\1{\bm{1}}

\DeclareMathAlphabet{\mathsfit}{\encodingdefault}{\sfdefault}{m}{sl}
\SetMathAlphabet{\mathsfit}{bold}{\encodingdefault}{\sfdefault}{bx}{n}

\usepackage{hyperref}
\usepackage{url}

\usepackage{verbatim}

\usepackage{algorithm}
\usepackage{algpseudocode}
\usepackage{graphicx}
\usepackage{tcolorbox}
\tcbuselibrary{skins, breakable}
\usepackage{pifont}
\usepackage{subcaption}
\usepackage{xcolor}
\usepackage[table]{xcolor}
\usepackage{threeparttable}
\usepackage{booktabs}
\usepackage{multirow}
\usepackage[utf8]{inputenc}
\usepackage{listings}
\usepackage{wrapfig}
\usepackage{float}
\usepackage{amsmath}
\usepackage[dvipsnames]{xcolor}

\definecolor{codegreen}{rgb}{0,0.6,0}
\definecolor{codegray}{rgb}{0.5,0.5,0.5}
\definecolor{codepurple}{rgb}{0.58,0,0.82}
\definecolor{backcolour}{rgb}{0.95,0.95,0.92}

\lstdefinestyle{mystyle}{
    backgroundcolor=\color{backcolour},   
    commentstyle=\color{codegreen},
    keywordstyle=\color{magenta},
    stringstyle=\color{codepurple},
    basicstyle=\ttfamily\footnotesize,
    breakatwhitespace=false,         
    breaklines=true,                 
    captionpos=b,                    
    keepspaces=true,                 
    showspaces=false,                
    showstringspaces=false,
    showtabs=false,                  
    tabsize=2
}

\lstdefinestyle{outputstyle}{
    basicstyle=\ttfamily\small,
    backgroundcolor=\color{backcolour},
}

\title{Reducing Cognitive Overhead in Tool Use via\\  Multi-Small-Agent Reinforcement Learning}

\author{
    \textbf{Dayu Wang}$^{1,2}$ \quad
    \textbf{Yutong Liu}$^{2}$ \quad
    \textbf{Jiaye Yang}$^{1}$ \\
    \textbf{Weikang Li}$^{3}$ \quad
    \textbf{Jiahui Liang}$^{1}$ \quad
    \textbf{Yang Li}$^{1}$ \\[3pt]
    $^{1}$Baidu Inc. \qquad
    $^{2}$Nanyang Technological University \qquad
    $^{3}$Peking University \\[3pt]
    \small
    \texttt{2100010872@stu.pku.edu.cn} \quad
    \texttt{c260089@e.ntu.edu.sg} \\
    \small
    \texttt{yamseyoung@gmail.com} \quad
    \texttt{wavejkd@pku.edu.cn} \\
    \small
    \texttt{\{liangjiahui03, liyang164\}@baidu.com}
}
\iclrfinalcopy
\begin{document}

\maketitle

\begin{abstract}
Recent progress in multi-agent systems highlights the promise of specialized agents that collaborate through a division of labor. In contrast, most tool-augmented reasoning systems still adopt a single-agent paradigm, where one large model must interleave high-level reasoning with fine-grained tool operations—a process that often leads to cognitive-load interference and unstable outputs. We propose \texttt{MSARL} (Multi-Small-Agent Reinforcement Learning), a novel framework that explicitly decouples reasoning from tool execution and interpretation. In \texttt{MSARL}, a dedicated reasoning agent focuses on strategic problem decomposition and planning, while a specialized tool agent processes long and complex tool outputs, acting as an adaptive condenser to bridge information gaps. This role-specific separation not only reduces cognitive interference but also accelerates the information flow. To enable effective collaboration, we introduce a hierarchical reinforcement learning approach that uses role-specific and collaboration-based rewards, providing granular feedback to the tool agent and a holistic, trajectory-level signal to the reasoning agent. On mathematical problem-solving with code execution, \texttt{MSARL} achieves more stable reasoning and higher final-answer accuracy than strong single-agent baselines. Our findings indicate that this dual-agent architecture significantly mitigates hallucinations and boosts tool invocation tendencies, thereby improving overall robustness. Our method provides a scalable blueprint for building specialized multi-agent system that can tackle complex reasoning tasks. The code for our method is available at: https://github.com/dayuwang401/MSARL-.
\end{abstract}

\section{Introduction}
\label{sec:introduction}

The emerging trend in agent-based AI systems is the specialization and collaboration of smaller, role-focused agents\citep{yang2023auto}.In tool-integrated reasoning, such division of labor promises gains in efficiency, interpretability, and scalability. Nevertheless, most existing systems still employ a single-agent paradigm, in which one large model sequentially performs high-level reasoning, generates executable tool calls (e.g., code), and interprets results\citep{xie2023openagents, gou2023tora, jin2025searchr1, qian2025toolrl,yao2023react}. While this integrated design simplifies coordination, it also introduces cognitive load interference: the same model must juggle long-horizon reasoning with precise, low-level tool operations.

We empirically examine this limitation by comparing a single, integrated agent to a decomposed one using identical model architectures and testing data. Despite having access to computational tools, the integrated agent produces fewer correct reasoning paths. This finding suggests that coupling high-level reasoning with tool execution in a single model can degrade the quality of intermediate logical steps.

Motivated by these observations, we present \texttt{MSARL} (Multi-Small-Agent Reinforcement Learning), a framework that decouples reasoning from tool use via explicit cognitive-role separation. In \texttt{MSARL}, a dedicated reasoning agent decomposes problems into stepwise plans and decides when to invoke tools, while other tool agents each specialize in a specific tool (e.g., code execution, retrieval API, calculator) to process lengthy and complex tool invocation information and results, compress the processed information, and pass it to the reasoning agent, thereby reducing the contextual complexity for the reasoning agent.

To enhance the collaboration capability among agents in handling specialized tasks, the agents are trained jointly through multi-agent reinforcement learning. In our multi-agent reinforcement learning framework, we introduce an innovative collaboration-oriented reward mechanism. Specifically, the reward received by one agent is determined by the quality of the output produced by another agent after processing the information it received from the former. The better the subsequent output of the collaborating agent, the higher the reward for the initiating agent at that step. This design grants agents substantial freedom to explore and discover optimal collaboration patterns for specialized tasks.

We first demonstrate \texttt{MSARL} on mathematical problem \citep{2025AIMEI,lei2024macm,MAA2023AMC,hendrycksmath2021,he-etal-2024-olympiadbench},solving via code execution, where it achieves higher reasoning stability and final-answer accuracy than single-agent baselines\citep{qwen2.5,zeng2025simplerl}. Beyond mathematics, the architecture naturally generalizes to multi-tool scenarios, offering a scalable blueprint for specialized-agent AI capable of tackling complex reasoning and decision-making tasks.


Our contributions can be summarized as follows:
\begin{itemize}
\item We conduct an in-depth empirical analysis of the limitations inherent in single-agent, tool-integrated reasoning systems, showing that coupling high-level reasoning and low-level tool execution within one model can degrade intermediate reasoning quality.
\item We propose \texttt{MSARL} (Multi-Small-Agent Reinforcement Learning), a novel framework that decouples cognitive roles via a dedicated Reasoning Agent and multiple specialized Tool Agents, equipped with a collaboration-oriented reward mechanism to optimize cooperation and information flow.
\item We validate the efficacy of \texttt{MSARL} through extensive experiments on mathematical problem solving and multi-tool reasoning tasks, demonstrating superior reasoning stability, final-answer accuracy, and scalability compared to single-agent baselines.
\end{itemize}

\section{Related Work}
\label{sec:related-work}
\paragraph{Tool-Integrated Reasoning.}
Tool-integrated reasoning (TIR) has emerged as a promising approach to enhance the capabilities of large language models (LLMs). By integrating external tools such as code interpreter~\citep{wang2023mathcoder, gou2023tora}, search engine~\citep{jin2025searchr1} or LLM-based agents~\citep{wu2025agentic}, TIR serves as an extension to a single executor, allowing models to perform more complex tasks. Despite its great potential, existing TIR approaches exhibit critical limitations. Previous studies distill trajectories from
stronger models and perform Supervised Fine-Tuning (SFT), limiting their ability to explore and adapt to optimal reasoning strategies. More recent research show the effectiveness of large-scale reinforcement learning (RL) for TIR with merely outcome rewards~\citep{openai2024o1, deepseekai2025deepseekr1, qwen3technicalreport}. Building on these advances, we focus on mathematical reasoning, a canonical domain for evaluating complex reasoning tasks, and generalize the notion of tool beyond traditional code interpreters to include any auxiliary agent that can support the reasoning process.

\paragraph{Multi-Agent System.} Leveraging multi-agent system (MAS) collaboration to complete complex tasks that are difficult to solve by single inference becomes increasingly popular~\citep{han2024llmmas, tran2025multi}. In mathematical reasoning, for example, \citet{yuan2025reinforce} propose an actor-critic architecture in which the critic generates multiple candidate answers and feedback to enable better self-reflection by the actor; \citet{zhang-xiong-2025-debate4math} introduce a debating paradigm with diverse agent roles to facilitate fine-grained reasoning through structured disagreement and adjudication. While predefined role interactions are widely adopted in MAS~\citep{lei2024macm, wang2024mixture, motwani2025malt}, some efforts have explored dynamic agent typology and interaction patterns, enabling more flexible and adaptive collaboration~\citep{zhuge2024language, zhang2025gdesigner, zhou2025reso}. However, compared to the success on the single LLM, existing MAS frameworks often lack reliable, fine-grained reward signals for MAS collaboration, relying instead on outputs or self-generated reward mechanisms. Falling into the paradigm of predefined agent collaboration, our work aims to enable the agents to learn how to interact with each other \textit{meanwhile} fine-tune their weights through feedback from cooperation.


\paragraph{Supervision Signals in RL.} Existing verification methods in RL can be categorized into three paradigms: outcome-based, process-based, and hybrid supervision. Outcome-based supervision evaluates final outcomes, as seen in the Outcome Reward Model (ORM)~\citep{ouyang2022training}. While simple, ORM lacks granularity for intermediate steps, resulting in reward sparsity. Process-based supervision addresses this by providing step-by-step or milestone feedback. Process-supervision Reward Models (PRMs)~\citep{lightman2023letsverifystepstep, qian2025toolrl} evaluates each reasoning step, enabling precise error correction. However, PRMs rely on costly human-annotated data, which might limit their scalability. Hybrid supervision integrates PRM’s step-level quality control with auxiliary rewards (e.g., Instruction Reward Model (IRM)~\citep{luo2024wizardmath}, which supervises quality of generated instruction) that supervise data quality from orthogonal angles (e.g., instruction quality). Although this complementary approach mitigates single-source limitations, it introduces optimization conflicts requiring careful calibration. These reward signals motivate diverse policy optimization methods such as PPO~\citep{schulman2017proximal}, DPO~\citep{rafailov2023direct}, GRPO~\citep{shao2024deepseekmath}. Among them, GRPO’s flexible reward function enables adaptation to diverse objectives, such as assigning weights to subtasks~\citep{yu2024steptool} or constraining tool use frequency~\citep{li2025torl}. In this work, we extend GRPO to enhance multi-agent collaboration in mathematical problem solving through a simple yet effective reward design.  
\section{Methodology}
\label{sec:method}
In this section, we start from a pilot investigation to see whether augmenting a single agent with code execution improves mathematical reasoning, or instead may introduce additional cognitive burden. Then we present the detailed design for training models by our proposal, i.e., \texttt{MSARL}, covering (1) rollout framework; (2) reward design; (3) training strategy.

\subsection{Cognitive Overhead in Single-Agent Reasoning}
\label{subsec:motivation}

\paragraph{Setup.} 
We conduct experiments on the non-reasoning model Qwen2.5-3B-Instruct~\citep{qwen2.5}, the reasoning model Qwen3-4B~\citep{qwen3technicalreport} and a mathmatical models Qwen2.5-Math-1.5B-Instruct~\citep{yang2024qwen25math} for mathmatical reasoning tasks. Specifically, two prompting regimes are considered with the same backbone model: (\romannumeral1) \textit{reasoning-only} (\texttt{r\_only}), in which the model is instructed to solve problems exclusively through step-by-step natural language reasoning; (\romannumeral2) \textit{reasoning with code} (\texttt{r\_code}), in which the model is encouraged to interleave Python code blocks with natural language reasoning, thereby allowing intermediate computations to be executed. Experiments are conducted on the MATH-500 test split~\citep{lightman2023lets}, a benchmark whose problems are naturally amenable to solution verification through executable code. For each problem, we draw $N=5$ independent samples per regime using nucleus sampling ($p=0.95$, temperature $=0.7$). To assess solution validity, we employ DeepSeek-R1~\citep{deepseekai2025deepseekr1} as the judge to evaluate whether a reasoning trajectory, disregarding minor arithmetic slips, constitutes a logically coherent path that would yield the correct answer under flawless execution. The complete prompts for two prompting strategies and model judge can be found in Appendix~\ref{app:motivation_prompt}.

\begin{figure*}[th]
\centering

\begin{subfigure}[t]{0.32\linewidth} 
\centering
\includegraphics[width=\linewidth]{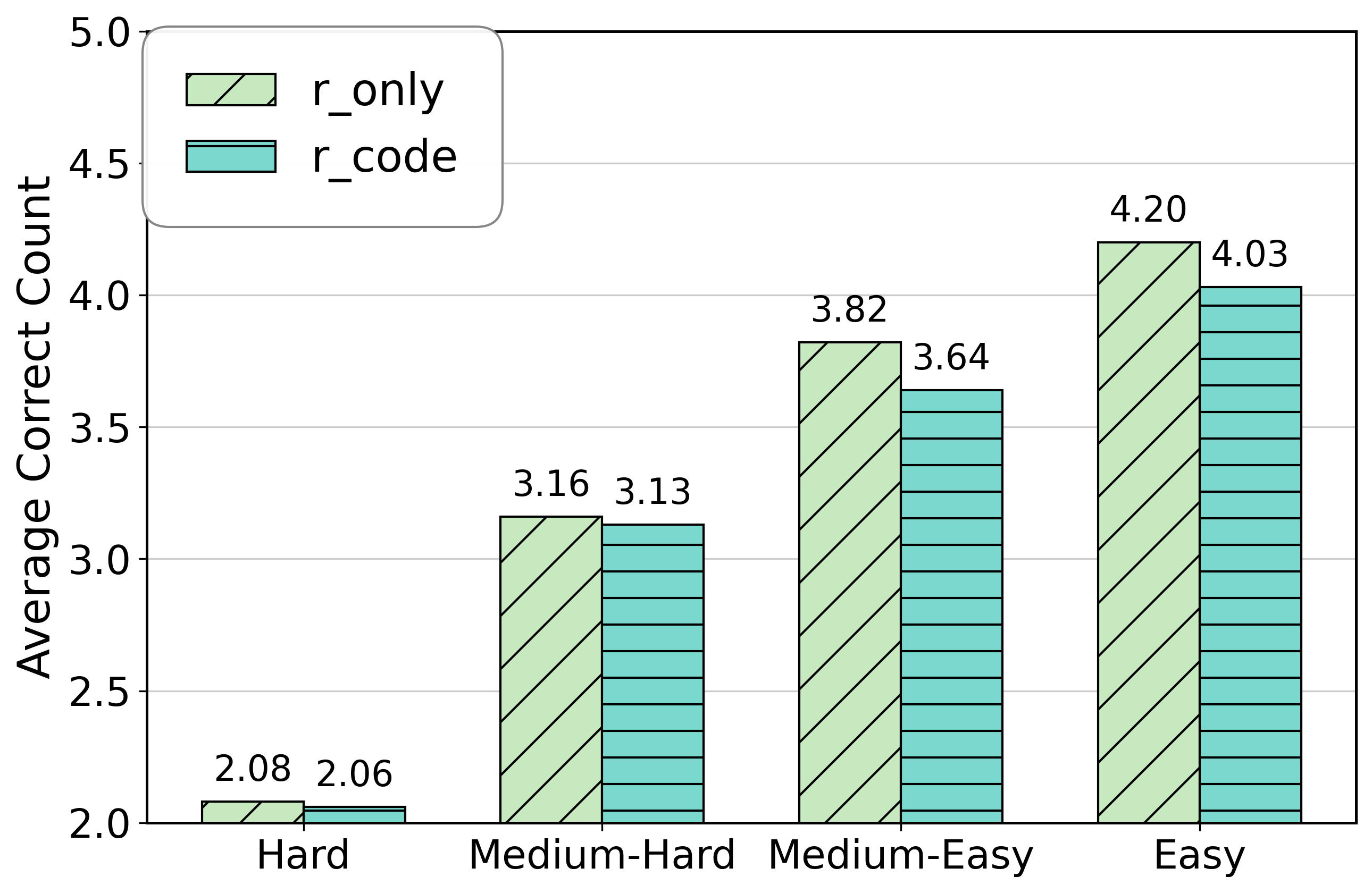}
\caption{Qwen2.5-3B-Instruct}
\label{fig:qwen2.5-3b-instruct-math500}
\end{subfigure}
\hspace{0em}
\begin{subfigure}[t]{0.32\linewidth}
\centering
\includegraphics[width=\linewidth]{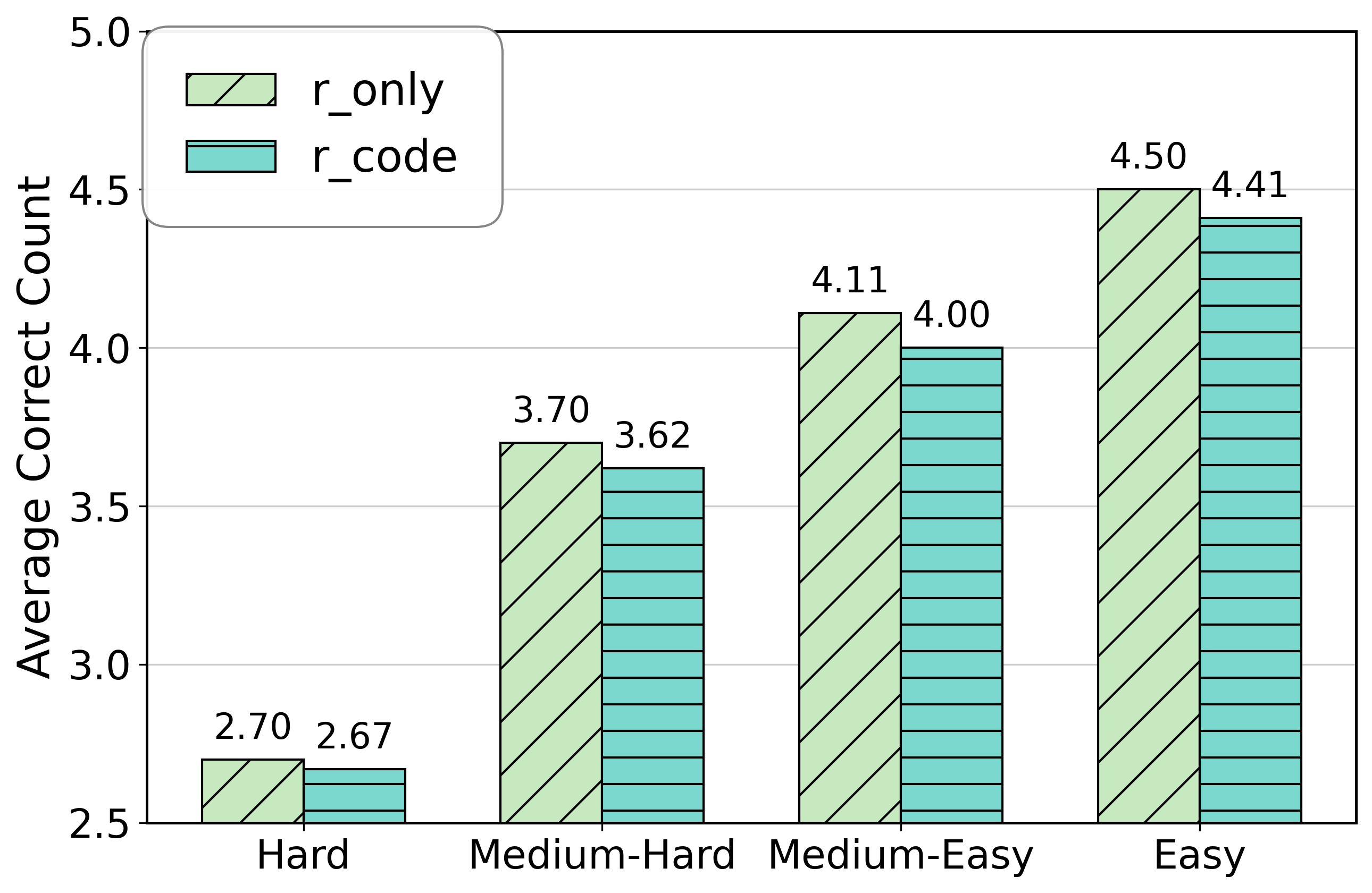}
\caption{Qwen2.5-Math-1.5B-Instruct}
\label{fig:qwen2.5-math-1.5b-math500}
\end{subfigure}
\hspace{0em}
\begin{subfigure}[t]{0.32\linewidth}
\centering
\includegraphics[width=\linewidth]{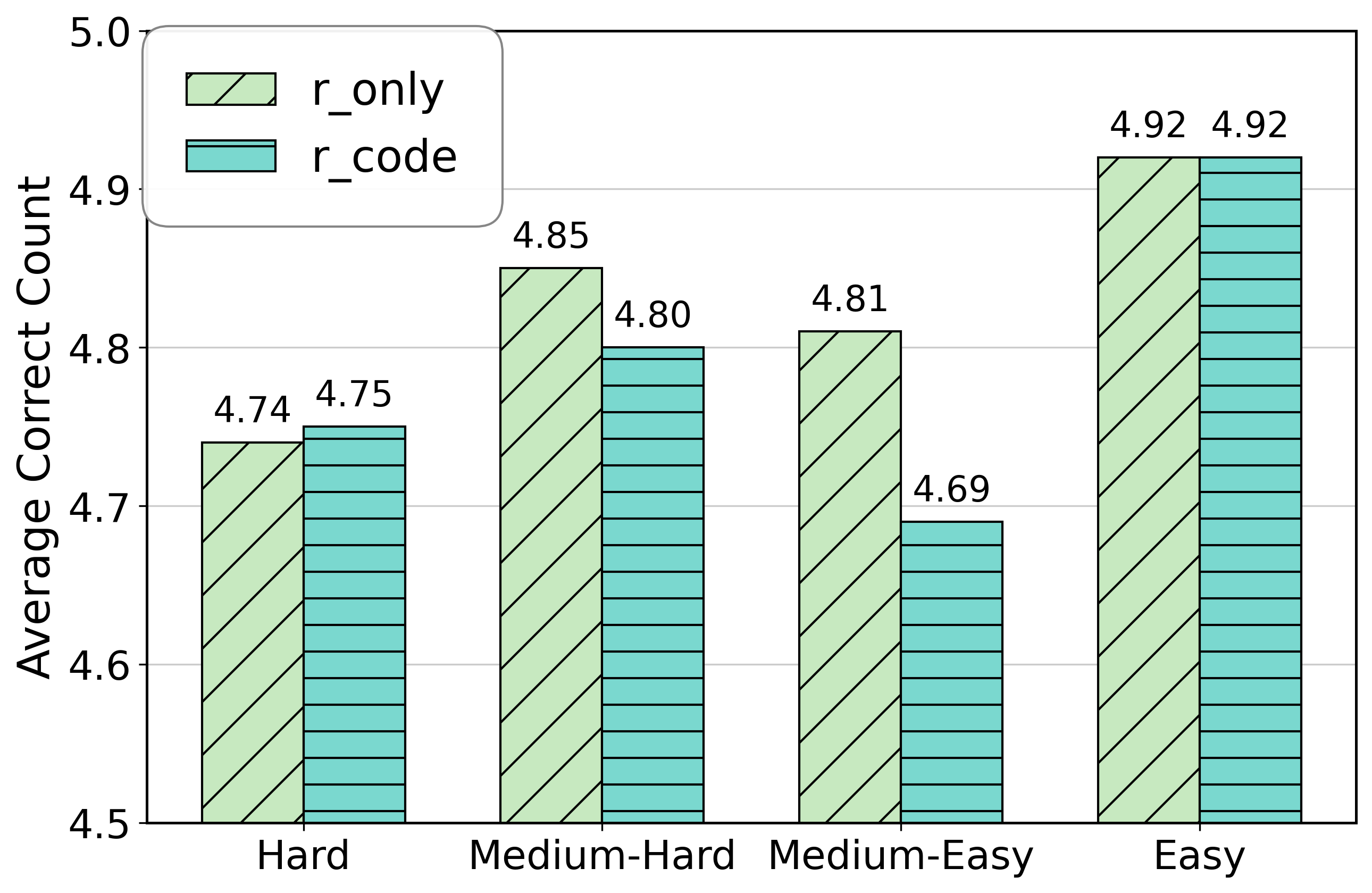}
\caption{Qwen3-4B}
\label{fig:qwen3-4b-math500}
\end{subfigure}

\caption{Model performance on Math-500 under two prompting regimes.}
\label{fig:motivation_result}
\end{figure*}

\paragraph{Results.} 
To analyze performance across different reasoning difficulty levels, we categorize MATH-500 problems into four groups based on their inherent difficulty levels (ranging from 1 to 5) and their respective proportions in the dataset\footnote{https://huggingface.co/datasets/HuggingFaceH4/MATH-500}: \texttt{Hard} (Level 5,~26.8\%), \texttt{Medium-Hard} (Level 4,~25.6\%), \texttt{Medium-Easy} (Level 3,~21\%), and \texttt{Easy} (Levels 1–2,~26.6\%).
As shown in Figure~\ref{fig:motivation_result}, the \texttt{r\_code} regime generally underperforms the \texttt{r\_only} regime, exhibiting accuracy gaps of $0.02$ to $0.18$ across all difficulty levels. The exception is Qwen3-4B, which demonstrates superior average performance on mathematical tasks, as expected given its strong reasoning capabilities in thinking mode~\citep{qwen3technicalreport}. The degradation is most pronounced on problems of medium-difficulty (\texttt{Medium-Hard}: $-0.08$, \texttt{Medium-Easy}: $-0.18$), while the performance gap is smaller on the more complex \texttt{Hard} problems. This pattern suggests that the additional cognitive load induced by generating, executing, and integrating code fragments is particularly disruptive when problems require a moderate level of reasoning complexity, where code execution might interfere with the initial problem formulation. To further illustrate this phenomenon, we provide qualitative analysis of a medium-difficulty example in Appendix~\ref{app:motivation_example}, contrasting model outputs under the two prompting regimes. These findings indicate that a single agent can struggle to effectively balance internal reasoning with external tool-use, suggesting a fundamental limitation in its ability to manage the cognitive load of both tasks simultaneously.

\subsection{Dual-Agent Framework for Decoupled Reasoning and Tool Use}
\begin{figure}[th]
\centering
\includegraphics[clip, width=1\linewidth]{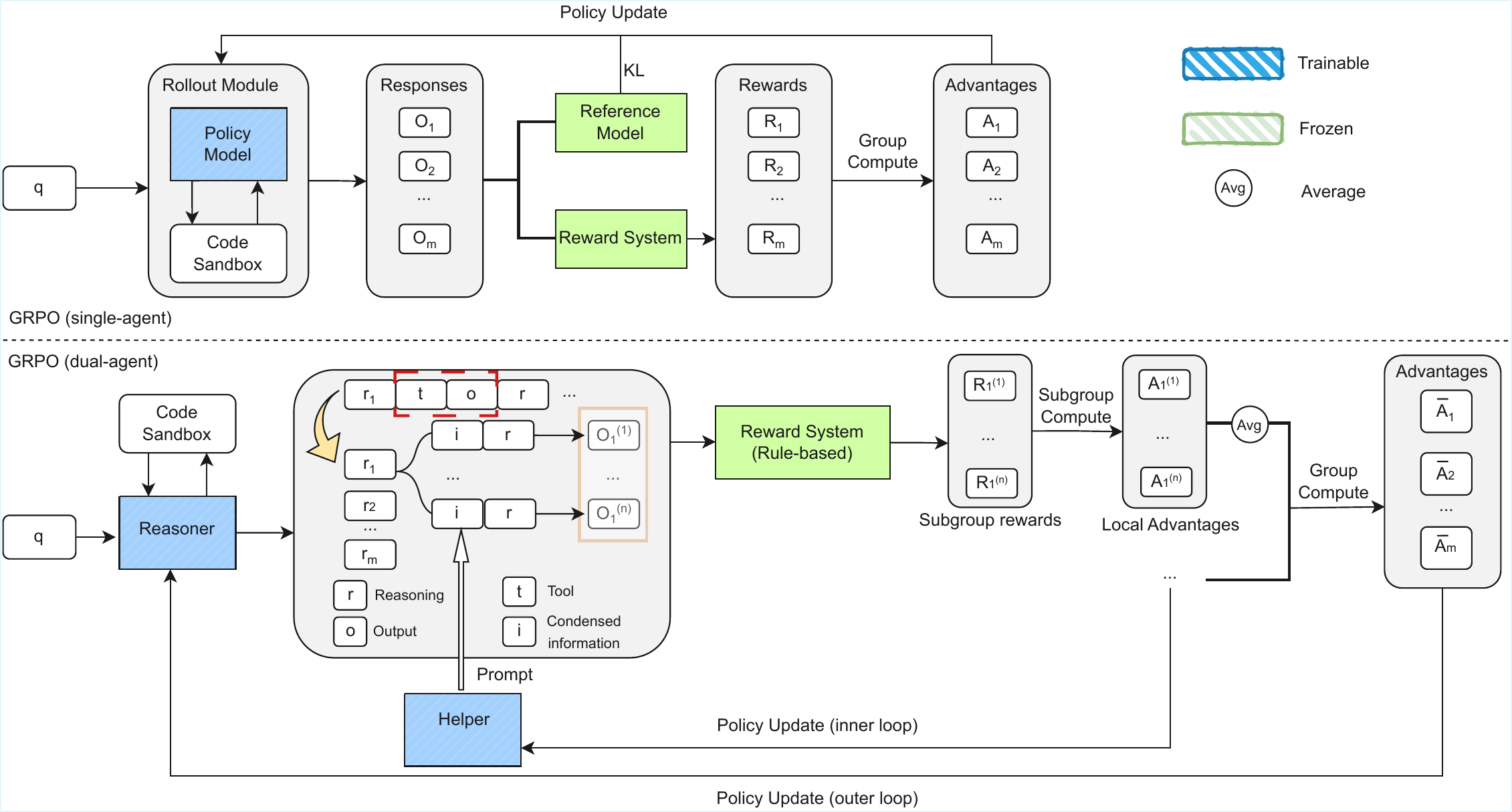} 
\caption{Overview of our method}
\label{fig:method_overview}
\end{figure}

\paragraph{Preliminaries and Notations.} Motivated by the observations in Section~\ref{subsec:motivation}, we propose \texttt{MSARL}, a \textit{dual-agent framework} designed to mitigate cognitive interference by explicitly decoupling the responsibilities of reasoning and tool-use. Figure~\ref{fig:method_overview} illustrates the framework of \texttt{MSARL}. In the proposed dual-agent system, one serves as a reasoning agent \emph{Reasoner} and the other as a tool agent \emph{Helper}. In the context of LLM policy optimization, let $\pi_{\text{reason}}$ be the reasoning LLM, $\pi_{\text{tool}}$ be the tool-assisting LLM. For each question $q$ in a given set $Q$, the \emph{Reasoner} generates one or more independent reasoning trajectories. Each reasoning trajectory consists of a sequence of steps $<r, t, o>$. At step $k$, the \emph{Reasoner} may produce either a natural language reasoning segment $r_k$ or a tool call $t_k$. When a tool call is executed, it returns the tool output $o_k$. A key step in our framework is the interaction with the \emph{Helper}. The pair $<t_k, o_k>$ is processed by the \emph{Helper} model $\pi_{\text{tool}}$ to generate a condensed, structured interpretation, denoted as $i_k$. This interpretation $i_k$ is then passed back to the \emph{Reasoner}, which uses the full context, including $r_k$ and $i_k$, to generate the subsequent reasoning step $r_{k+1}$. This iterative process continues until the \emph{Reasoner} produces a final answer, denoted as $O$. 

\begin{figure}[htbp]
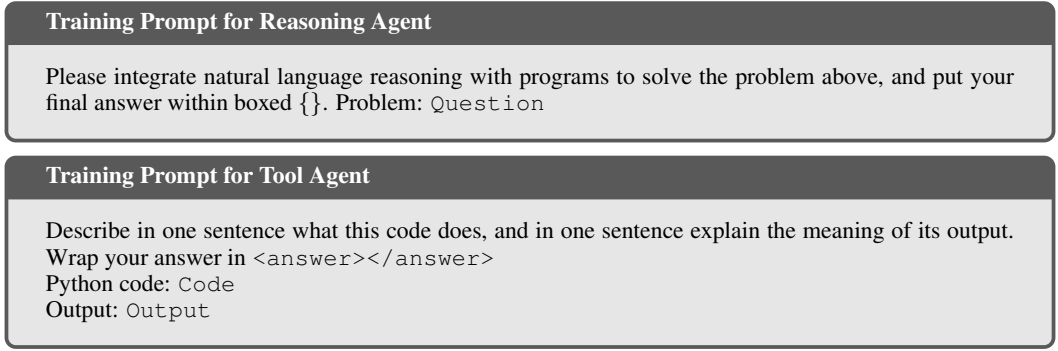

\centering
\footnotesize 

\begin{tcolorbox}[
  colframe=gray!70!black,  
  colback=black!10,        
  title=Training Prompt for Reasoning Agent,
  coltitle=white,
  fonttitle=\bfseries
]
\small 
Please integrate natural language reasoning with programs to solve the problem above, and put your final answer within boxed \{\}. Problem: \texttt{Question}
\end{tcolorbox}

\begin{tcolorbox}[
  colframe=gray!70!black,  
  colback=black!10,        
  title=Training Prompt for Tool Agent,
  coltitle=white,
  fonttitle=\bfseries
]
\small 
Describe in one sentence what this code does, and in one sentence explain the meaning of its output. Wrap your answer in \texttt{<answer></answer>}

Python code: \texttt{Code}

Output: \texttt{Output}
\end{tcolorbox}

\caption{Training prompt templates for \texttt{MSARL}}
\label{fig:training-prompts}
\end{figure}

\paragraph{Rollout Framework.} To enable the model to autonomously generate reasoning traces and tool calls, we utilize the prompts as shown in Figure~\ref{fig:training-prompts}. Following~\cite{li2025torl}, when a code termination identifier (\texttt{```output}) is detected, the \emph{Reasoner} pauses generation; the latest code block is extracted for code execution in the code sandbox; both the code \texttt{Code} and the structured execution result \texttt{Output} are wrapped by training prompt for tool agent, resulting in \emph{Helper}'s output inserted into the \texttt{<answer>} field. The \emph{Reasoner}'s final answer is indicated within a box (e.g., $\boxed{final\_answer}$). We observed the agent-to-agent communication during the rollout process introduces significant GPU idle time. To maximize the training efficiency, we introduced a hyperparameter, $C$, which is the maximum number of tool calls the model can make during a single response generation.  Once this threshold is exceeded, the system ignores further code execution requests, forcing the model to switch to pure-text reasoning mode. To better illustrate the training dynamics, we provide a full rollout example in Appendix~\ref{app:method_example}. 

\paragraph{Reward Design.} A critical challenge in optimizing multi-step reasoning agents is the sparsity of terminal reward signals, which often fail to provide sufficient guidance for intermediate reasoning steps. To address this issue, our reward design strategically applies the outcome-based reward at the most decisive juncture, that is, the tool-use interpretation stage as identified in Section~\ref{subsec:motivation}. Specifically, the binary success reward (Eq.~\ref{eq:reward_model}) is granted based on the output of the \emph{Helper} agent ($\pi_{\text{tool}}$). It checks whether the \emph{Reasoner} model output is both calculated and formatted correctly, as specified by the ground truth. This approach reframes the final outcome evaluation as a \textit{dense and immediate milestone reward} for effective tool utilization. Crucially, during joint optimization, this reward signal propagates back to the \emph{Reasoner} policy ($\pi_{\text{reason}}$). This process operates as an implicit regularization mechanism, incentivizing the \emph{Reasoner} to generate logically sound reasoning traces that culminate in correct and executable tool invocations.  

\begin{equation}
\mathcal{R} = 
\begin{cases}
1, & \parbox{5.5cm}{if the answer is correct and boxed} \vspace{8pt}\\
0, & \text{otherwise}
\end{cases}
\label{eq:reward_model}
\end{equation}

\paragraph{\emph{Reasoner}-\emph{Helper} Framework.} Our methodology is built upon a hierarchical reinforcement learning framework designed to jointly optimize both the reasoning policy ($\pi_{\text{reason}}$) and the tool policy ($\pi_{\text{tool}}$). 
To tune the model with structured rewards, we employ GRPO~\citep{shao2024deepseekmath} that introduces advantage normalization within grouped samples. For brevity, assume that $\pi_{\text{reason}}$ parallelly generates $m$ reasoning trajectories up to the point of its final tool invocation from the question $q$. For such a specific clipped reasoning trajectory $\tau_k$ containing tool execution results. At this juncture, the \emph{Helper} policy $\pi_{\text{tool}}$ generates a set of $n$ diverse interpretations of the tool's output, denoted as $\{i_k^{(1)}, i_k^{(2)}, \ldots, i_k^{(n)}\}$. Each of these interpretations is then passed back to the \emph{Reasoner}. The \emph{Reasoner} acts as a proxy to complete the reasoning process, generating a final conclusive segment for each interpretation. This branching process results in a set of $n$ distinct final answers, $\{O_k^{(1)}, O_k^{(2)}, \ldots, O_k^{(n)}\}$. Next, the reward system in Eq.~\ref{eq:reward_model} scores each output $O_k^{(j)}$ based on its correctness relative to the ground truth ($GT$), yielding a reward value $R_k^{(j)}$. Although the reward is computed on the final answer $O_k^{(j)}$, it acts as a direct performance signal for the corresponding interpretation $i_k^{(j)}$ that produced it. These outputs and their corresponding reward values form an evaluation subgroup $G_k$:
$$
G_k = \{(O_k^{(1)}, R_k^{(1)}), (O_k^{(2)}, R_k^{(2)}), \ldots, (O_k^{(n)}, R_k^{(n)}) \}
$$

We then calculate the normalized advantage $A_k^{(j)}$ for each output relative to the mean $\mu_k$  and standard deviation $\sigma_k$ of rewards within the subgroup $G_k$. A small constant $\eta$ is added for numerical stability. The normalized advantage for each output is then defined as:
\begin{equation}
    A_k^{(j)} = \frac{R_k^{(j)} - \mu_k}{\sigma_k + \eta}
\end{equation}

\begin{small}
\begin{equation}
    \mu_k = \frac{1}{n} \sum_{j=1}^n R_k^{(j)}, \quad \sigma_k = \sqrt{\frac{1}{n} \sum_{j=1}^n (R_k^{(j)} - \mu_k)^2}
\end{equation}
\end{small}

Since the ultimate goal is to optimize the generation of the entire reasoning trajectory, we perform average pooling on the advantage values computed in the clipped reasoning trajectory $\tau_k$ to obtain an aggregated advantage value $\bar{A}_k$ that represents the overall quality of the entire trajectory:
\begin{equation}
    \bar{A}_k = \frac{1}{n} \sum_{j=1}^n A_k^{(j)}
\end{equation}

The training objectives, adapted from GRPO, are designed to separately update the parameters of the tool policy ($\theta_{\text{tool}}$) and the reasoning policy ($\theta_{\text{reason}}$). First, we optimize the tool policy $\pi_{\text{tool}}$. The goal is to encourage the \emph{Helper} agent to generate interpretations that lead to correct final answers. The normalized advantage $A_k^{(j)}$ is applied at the token level to the corresponding interpretation $i_k^{(j)}$ that leads to the final outcome $O_k^{(j)}$. The learning objective of $\pi_{\text{tool}}$ is formulated as:

\begin{equation}
\label{eq:obj_tool}
\begin{aligned}
\mathcal{J}(\theta_{\text{tool}}) =
\mathbb{E}{\tau_k, {i_k^{(j)}}} \\
&
\left[
\frac{1}{n} \sum_{j=1}^n \frac{1}{|i_k^{(j)}|} \sum_{t=1}^{|i_k^{(j)}|}
\min \left(
r_{t}^{(j)}(\theta_{\text{tool}}) A_k^{(j)},
\text{clip}(r_{t}^{(j)}(\theta_{\text{tool}}), 1 - \epsilon, 1 + \epsilon) A_k^{(j)}
\right)
\right]
\end{aligned}
\end{equation}
where $r_{t}^{(j)}(\theta_{\text{tool}})= \frac{\pi_{\text{tool},\theta}(i_{k, t}^{(j)} | \tau_k, i_{k,<t}^{(i)})}{\pi_{\text{tool,old}}(i_{k,t}^{(j)} | \tau_k, i_{k,<t}^{(j)})}$ is the probability ratio for the $t$-th token of the $j$-th interpretation, and the expectation is over the distribution of trajectories and interpretations.

Next, we optimize the reasoning policy $\pi_{\text{reason}}$. The objective is to guide the \emph{Reasoner} to generate trajectories that are more likely to result in high-reward outcomes after tool interpretation. The aggregated advantage $\bar{A}_k$, representing the average quality of a reasoning path $\tau_k$, is applied to every token within that path. The objective is thus defined as:
\begin{equation}
\label{eq:obj_reason}
\begin{aligned}
\mathcal{J}(\theta_{\text{reason}}) =
\mathbb{E}{{\tau_k}}
\left[
\frac{1}{m} \sum_{k=1}^m \frac{1}{|\tau_k|} \sum_{t=1}^{|\tau_k|}
\min \left(
r_{t}^{(k)}(\theta_{\text{reason}}) \bar{A}_k,
\text{clip}(r_{t}^{(k)}(\theta_{\text{reason}}), 1 - \epsilon, 1 + \epsilon) \bar{A}_k
\right)
\right]
\end{aligned}
\end{equation}
where $r{t}^{(k)}(\theta_{\text{reason}}) = \frac{\pi_{\text{reason},\theta}(\tau_{k,t} | q, \tau_{k,<t})}{\pi_{\text{reason,old}}(\tau_{k,t} | q, \tau_{k,<t})}$ is the probability ratio for the $t$-th token of the $k$-th reasoning trajectory.


Recent work~\citep{qian2025toolrl, zhang2025critiquegrpo} suggest that the omission of KL penalty term against a reference model can encourage the model to more freely adapt its behavior to our custom response format and structured reward signals. Following their implementation, we remove the term to simplify the training pipeline while gain comparable performance in practice. We summarize the process for advantage estimation and policy update  in Appendix~\ref{app:refinement_process}.

\section{Experiments}
\label{sec:experiments}

\subsection{IMPLEMENTATION DETAILS}
\paragraph{Datasets and Metrics.} The training utilizes approximately 1.2w queries from all difficulty levels of the MATH training dataset~\citep{hendrycksmath2021}. We evaluate the proposed method on widely used mathmatical benchmarks, including AIME 2024~\citep{li2024numinamath}, AIME25~\citep{2025AIMEI}\footnote{https://modelscope.cn/datasets/TIGER-Lab/AIME25}, MATH-500~\citep{hendrycksmath2021}, OlympiadBench~\citep{he-etal-2024-olympiadbench}, AMC23~\citep{MAA2023AMC}. We report Pass@1 as in~\citep{cui2025prime}. In addition, to measure stability and consistency of model generated responses, we include Pass@8 and Maj@8 on all benchmarks, following prior work~\citep{yang2024qwen25math, zhang2025lessons}.

\paragraph{Models and Baselines.} We choose Qwen2.5-Math-Instruct~\citep{yang2024qwen25math} series models as backbones for their superior mathmatical capabilities. Specifically, we use Qwen2.5-Math-1.5B-Instruct as the backbone of the reasoning agent, and Qwen2.5-1.5B-Instruct as the backbone of the tool agent.  We compare our method mainly against reinforcement learning-based approaches, most of which are single-agent architecture.

\paragraph{Training and Evaluation Details.}  We use Sandbox Fusion as the code interpreter during training and evaluation. Due to the dual-agent framework, the global batch size is set to 1 by default. The \emph{Reasoner} generates 3 samples per question and the \emph{Helper} uses a rollout size of 3. To maximize training efficiency, the default maximum number of tool calls $C$ is set to 1. All models are RL-tuned with a cold start. For evaluation, we use greedy decoding (temperature = 0) across all models. For a fair comparison, we set the maximum number of tool calls to 1 during inference.

\subsection{Main Results}
$\triangleright$ \texttt{MSARL}-1.5B achieves the highest average pass@1 accuracy of $55.9\%$ across all five datasets, as shown in Table ~\ref{table: main-result}. This represents a substantial improvement of $5.9\%$ over the strongest baseline, Qwen2.5-Math-1.5B-Instruct-TIR ($50.0\%$). \texttt{MSARL}-1.5B also surpasses other methods on challenging MATH500 ($77.6\%$) and Olympiad ($49.0\%$) datasets. This clear advantage confirms the effectiveness of introducing agents joint learning into a multi-agent system. Furthermore, our method show remarkable parameter efficiency. Our 1.5B models surpasses most competing 7B models by a considerable margin.  For instance, it outperforms SimpleRL-Zero and Qwen2.5-Math-7B-Instruct by ${14.2, 16.1}$ in pass@1 accuracy. This result implies that our \texttt{MSARL} approach provides a more effective and efficient path to enhancing mathematical reasoning in language models than simply scaling up model size or applying standard fine-tuning techniques.

\begin{table*}[htbp]
\centering
\caption{Comparison of different models testing accuracy on mathematical benchmarks with Pass@1. The best two performance are \textbf{bold} and \underline{underlined}.}
\label{table: main-result}
\resizebox{1.0\textwidth}{!}{
    \begin{tabular}{lcccccc}
        \toprule
        Model & AIME24 & AIME25 & MATH500 & Olympiad & AMC23 & Avg \\ 
        \midrule
        \multicolumn{6}{c}{\textit{Models based on Qwen2.5-Math-1.5B-Base}} \\
        Qwen2.5-Math-1.5B-Instruct & 10.0 & 10.0 & 66.0 & 31.0 & 62.5 & 35.9\\
        Qwen2.5-Math-1.5B-Instruct-TIR & \underline{23.3} & \textbf{20.0} & \underline{75.6} & \underline{48.5} & \underline{62.5} & \underline{50.0} \\ 
        \midrule
        \multicolumn{6}{c}{\textit{Models based on Qwen2.5-Math-7B-Base}} \\
        Qwen2.5-Math-7B-Instruct & 10.0 & 16.7 & 74.8 & 32.4 & \textbf{65.0} & 39.8\\
        Qwen2.5-Math-7B-Instruct-TIR & 20.0 & 6.7 & 70.4 & 45.0 & 50.0 & 34.2\\
        SimpleRL-Zero & \textbf{30.0} & \underline{20.0} & 66.8 & 29.0 & 62.5 & 41.7\\ 
        Eurus-2-7B-PRIME & 10.0 & 13.3 & 62.8 & 42.1 & 50.0 & 35.6\\ 
        \texttt{MSARL}-1.5B (Untrained) & 23.3 & 20.0 & 74.4 & 47.0 & 62.5 & \underline{52.6}\\ 
        \rowcolor{gray!20} \texttt{MSARL}-1.5B (Ours) & 
$16.7$ & 
$16.7$ & 
\textbf{$77.6$} & 
\textbf{$49.0$} & 
$57.5$ & 
\textbf{$55.9_{\textcolor{ForestGreen}{+5.9}}$}  \\ 
        \bottomrule
    \end{tabular}
}
\end{table*}

$\triangleright$Our method generally shows strong performance, securing either the best or second-best results in a majority of the pass@8 and maj@ metrics as summarized in Table~\ref{table: main-result-pass8-maj8}. Specifically, it achieves the best maj@8 score on MATH500 and the top scores for both pass@8 and maj@8 on the Olympiad dataset. It also ties for the best pass@8 on AMC23 and has the highest maj@8 score on that same dataset. As the maj@8 metric is a more robust indicator of a model's consistency and confidence, \texttt{MSARL} strong performance on maj@8 across most datasets suggests that its dual-agent approach leads to more reliable and stable outputs.

\begin{table*}[htbp]
\centering
\caption{Comparison of different models on mathematical benchmarks with Pass@8 and Maj@8. The best two performance are \textbf{bold} and \underline{underlined}.}
\label{table: main-result-pass8-maj8}
\resizebox{1.0\textwidth}{!}{
    \begin{tabular}{l|cc|cc|cc|cc|cc}
        \toprule
        Model & 
        \multicolumn{2}{c|}{AIME24} & 
        \multicolumn{2}{c|}{AIME25} & 
        \multicolumn{2}{c|}{MATH500} & 
        \multicolumn{2}{c|}{Olympiad} & 
        \multicolumn{2}{c}{AMC23} \\
        & pass@8 & maj@8 & pass@8 & maj@8 & pass@8 & maj@8 & pass@8 & maj@8 & pass@8 & maj@8 \\
        \midrule
        Qwen2.5-Math-1.5B-Instruct-TIR & 40.0 & 20.0 & \textbf{40.0} & \textbf{30.0} & \textbf{93.4} & 81.6 & 66.5 & 54.3 & \underline{87.5} & 60 \\
        Qwen2.5-Math-7B-Instruct-TIR & \underline{46.7} & \underline{26.7} & 26.7 & 13.3 & 88.8 & 78.4 & 63.9 & 48.1 & 80 & \underline{67.5} \\
        SimpleRL-Zero & \textbf{50.0} & \textbf{30.0} & 26.7 & 20.0 & 90.2 & \underline{82.0} & - & - & 85.0 & 67.5 \\
        Eurus-2-7B-PRIME & 46.7 & 20.0 & 36.7 & 16.7 & 90.2 & 73.4 & 60.4 & 40.2 & 85.0 & 57.5 \\
        \rowcolor{gray!20}\texttt{MSARL}-1.5B (Ours) & 40.0 & 23.3 & \underline{36.7} & \underline{20.0} & \underline{92} & \textbf{82.6} & \textbf{77.3} & \textbf{54.7} & \textbf{87.5}& \textbf{72.5}\\ 
        \bottomrule
    \end{tabular}
}
\end{table*}

\subsection{ABLATION STUDY AND EXTRA INVESTIGATION}
\paragraph{Contribution of Joint Training.} To verify that the superior performance of our method originates from the proposed joint optimization strategy, rather than merely from the inherent structure of a dual-agent workflow, we replace our two trained agents with their respective untrained models and evaluate this configuration. A performance degradation $-3.6$ in the untrained setup is observed in the penultimate row in Table~\ref{table: main-result}, which confirms the effectiveness of our joint optimization framework. Interestingly, we observe that even the dual-agent architecture with untrained models surpasses the strongest single LLM baseline in a zero-shot setting.   This validates the effectiveness of the decoupling strategy on complex reasoning problems in its own right.

\paragraph{Evaluation across Different Training Steps.}

\begin{wrapfigure}{r}{0.5\textwidth}
    \vspace{-10pt} 
    \centering
    \includegraphics[width=1.0\linewidth]{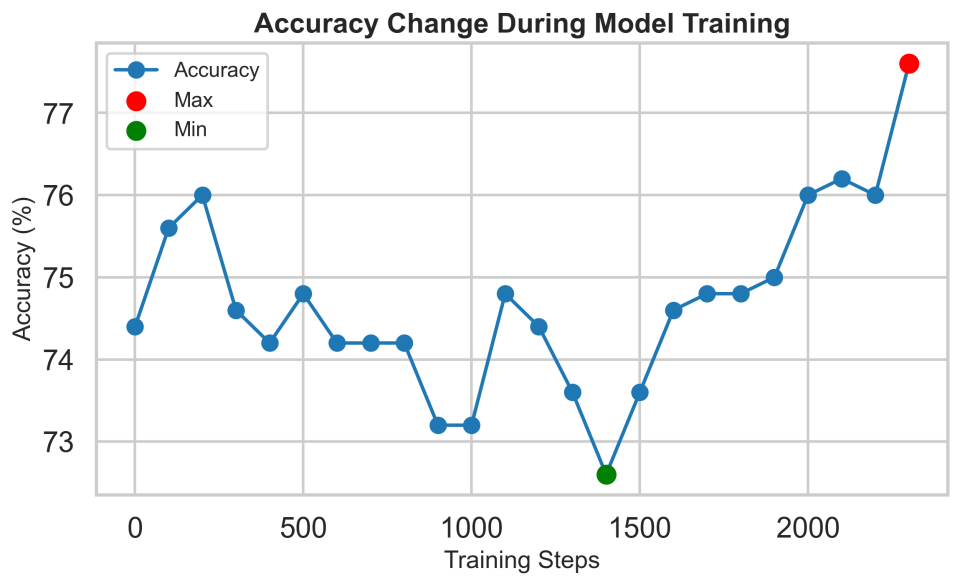}
    \caption{The model's performance (Average Pass@1) at different training checkpoints. Performance saturates after 2k steps.}
    \label{fig:training_steps}
    \vspace{-15pt} 
\end{wrapfigure}
To analyze the training dynamics and convergence of our framework, we evaluated its performance at various training checkpoints. Figure~\ref{fig:training_steps} plots the average Pass@1 accuracy on our validation benchmarks as a function of the number of training steps.
Figure~\ref{fig:training_steps} reveals a rapid performance improvement during the early stages of training, which is followed by a steady but diminishing rate of gain as training progresses. The frameworks's performance begins to saturate around the 2k training step mark, indicating stable convergence without signs of overfitting on the test data. This analysis shows the robustness of our training process and justifies our selection of the final model checkpoint for the main evaluation.

\paragraph{Reward Dynamics During Training.}
\begin{wrapfigure}{r}{0.5\textwidth}
    \vspace{-10pt} 
    \centering
    \includegraphics[width=1.0\linewidth]{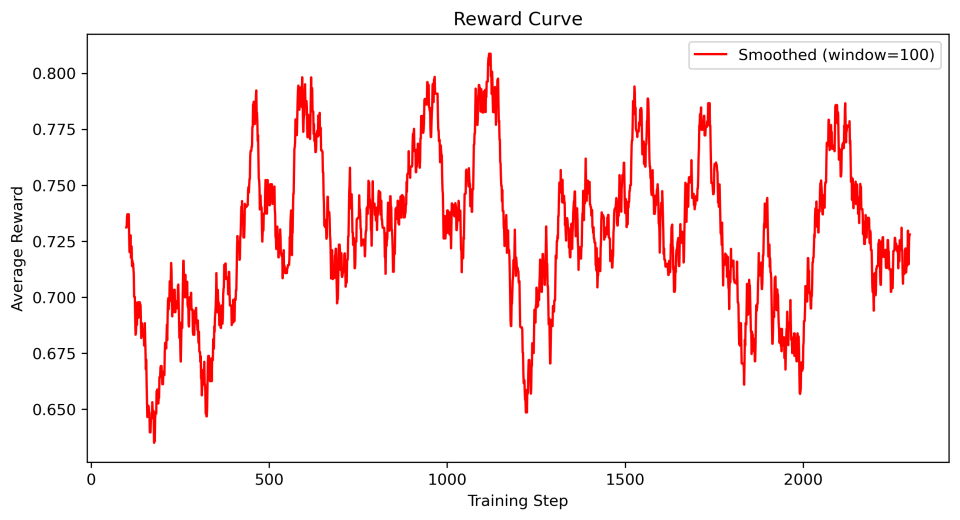}
    \caption{Average reward score during training. The consistent upward trend demonstrates successful and stable learning, with the policy converging in the later stages.}
    \label{fig:reward_dynamics}
    \vspace{-15pt} 
\end{wrapfigure}
To provide insight into the learning process of our framework, we analyze the reward dynamics during training. Figure~\ref{fig:reward_dynamics} plots the average reward score obtained by the agents as a function of training steps. The curve, smoothed with a moving average for clarity, first shows a stable increase indicating that our agent consistently improves its policy to maximize rewards. The absence of significant volatility or collapses in the reward curve confirms the robustness of our training configuration and hyperparameter settings. The reward begins to plateau in the later stages, suggesting that the agent is converging to an effective and stable policy. 
\section{Conclusion}
\label{sec:conclusion}
In this paper, we first identify a key challenge: the cognitive load interference inherent in the single-agent paradigm for tool-integrated reasoning. We then propose \texttt{MSARL}, a novel dual-agent framework that explicitly decouples high-level reasoning from low-level tool interpretation. This division of labor, powered by two specialized agents, significantly enhances the efficiency of information flow and addresses the identified interference. Specifically, \texttt{MSARL} leverages a reasoning agent for macro-level problem planning and a tool agent for micro-level, adaptive tool output interpretation. To optimize these agents jointly, we introduce a hierarchical reinforcement learning approach based on GRPO. This method uses normalized advantages to provide a granular, high-fidelity training signal to the tool agent and aggregated advantages to provide a consistent, trajectory-level signal to the reasoning agent. Extensive experiments on mathematical problem-solving tasks demonstrate that \texttt{MSARL} consistently achieves superior performance and higher reasoning stability compared to single-agent baselines. Furthermore, the modular nature of \texttt{MSARL} naturally generalizes to multi-tool scenarios, offering a scalable blueprint for future agent-based systems. Future work could explore extending \texttt{MSARL} to scenarios with more diverse and complex tool sets and information sources.

\clearpage
\bibliography{iclr2025_conference}

@article{zeng2025simplerl,
  title={Simplerl-zoo: Investigating and taming zero reinforcement learning for open base models in the wild},
  author={Zeng, Weihao and Huang, Yuzhen and Liu, Qian and Liu, Wei and He, Keqing and Ma, Zejun and He, Junxian},
  journal={arXiv preprint arXiv:2503.18892},
  year={2025}
}

@inproceedings{yao2023react,
  title={React: Synergizing reasoning and acting in language models},
  author={Yao, Shunyu and Zhao, Jeffrey and Yu, Dian and Du, Nan and Shafran, Izhak and Narasimhan, Karthik and Cao, Yuan},
  booktitle={International Conference on Learning Representations (ICLR)},
  year={2023}
}

@article{xie2023openagents,
  title={Openagents: An open platform for language agents in the wild},
  author={Xie, Tianbao and Zhou, Fan and Cheng, Zhoujun and Shi, Peng and Weng, Luoxuan and Liu, Yitao and Hua, Toh Jing and Zhao, Junning and Liu, Qian and Liu, Che and others},
  journal={arXiv preprint arXiv:2310.10634},
  year={2023}
}

@article{yang2023auto,
  title={Auto-gpt for online decision making: Benchmarks and additional opinions},
  author={Yang, Hui and Yue, Sifu and He, Yunzhong},
  journal={arXiv preprint arXiv:2306.02224},
  year={2023}
}

@misc{qwen2.5,
    title = {Qwen2.5: A Party of Foundation Models},
    url = {https://qwenlm.github.io/blog/qwen2.5/},
    author = {{Qwen Team}},
    month = {September},
    year = {2024}
}

@misc{deepseekai2025deepseekr1,
      title={DeepSeek-R1: Incentivizing Reasoning Capability in LLMs via Reinforcement Learning}, 
      author={DeepSeek-AI},
      year={2025},
      eprint={2501.12948},
      archivePrefix={arXiv},
      primaryClass={cs.CL},
      url={https://arxiv.org/abs/2501.12948}, 
}

@misc{qwen3technicalreport,
      title={Qwen3 Technical Report}, 
      author={{Qwen Team}},
      year={2025},
      eprint={2505.09388},
      archivePrefix={arXiv},
      primaryClass={cs.CL},
      url={https://arxiv.org/abs/2505.09388}, 
}

@article{yang2024qwen25math,
  title={Qwen2.5-Math Technical Report: Toward Mathematical Expert Model via Self-Improvement}, 
  author={An Yang and Beichen Zhang and Binyuan Hui and Bofei Gao and Bowen Yu and Chengpeng Li and Dayiheng Liu and Jianhong Tu and Jingren Zhou and Junyang Lin and Keming Lu and Mingfeng Xue and Runji Lin and Tianyu Liu and Xingzhang Ren and Zhenru Zhang},
  journal={arXiv preprint arXiv:2409.12122},
  year={2024}
}

@article{lightman2023lets,
      title={Let's Verify Step by Step}, 
      author={Lightman, Hunter and Kosaraju, Vineet and Burda, Yura and Edwards, Harri and Baker, Bowen and Lee, Teddy and Leike, Jan and Schulman, John and Sutskever, Ilya and Cobbe, Karl},
      journal={arXiv preprint arXiv:2305.20050},
      year={2023}
}

@article{hendrycksmath2021,
  title={Measuring Mathematical Problem Solving With the MATH Dataset},
  author={Dan Hendrycks and Collin Burns and Saurav Kadavath and Akul Arora and Steven Basart and Eric Tang and Dawn Song and Jacob Steinhardt},
  journal={NeurIPS},
  year={2021}
}

@article{li2024numinamath,
  title={Numinamath: The largest public dataset in ai4maths with 860k pairs of competition math problems and solutions},
  author={Li, Jia and Beeching, Edward and Tunstall, Lewis and Lipkin, Ben and Soletskyi, Roman and Huang, Shengyi and Rasul, Kashif and Yu, Longhui and Jiang, Albert Q and Shen, Ziju and others},
  journal={Hugging Face repository},
  volume={13},
  pages={9},
  year={2024}
}

@misc{2025AIMEI,
  author       = {{MAA}},
  title        = {2025 {AIME} I Problems},
  year         = {2025},
  note         = {Accessed: 2025-09-17},
  url          = {https://artofproblemsolving.com/wiki/index.php/2025_AIME_I?srsltid=AfmBOoof5gaaqlt3-l6LH7Tt6qmJZtl_2PQEDYlLFlMqhq9dLL8FMCRR}
}

@inproceedings{he-etal-2024-olympiadbench,
    title = "{O}lympiad{B}ench: A Challenging Benchmark for Promoting {AGI} with Olympiad-Level Bilingual Multimodal Scientific Problems",
    author = "He, Chaoqun  and Luo, Renjie  and Bai, Yuzhuo  and Hu, Shengding  andThai, Zhen  and Shen, Junhao  and Hu, Jinyi  and Han, Xu  and Huang, Yujie  and Zhang, Yuxiang  and Liu, Jie  and Qi, Lei  and Liu, Zhiyuan  and Sun, Maosong",
    editor = "Ku, Lun-Wei  and Martins, Andre  and Srikumar, Vivek",
    booktitle = "Proceedings of the 62nd Annual Meeting of the Association for Computational Linguistics (Volume 1: Long Papers)",
    month = aug,
    year = "2024",
    address = "Bangkok, Thailand",
    publisher = "Association for Computational Linguistics",
    url = "https://aclanthology.org/2024.acl-long.211/",
    doi = "10.18653/v1/2024.acl-long.211",
    pages = "3828--3850",
}

@misc{MAA2023AMC,
  author       = {{MAA}},
  title        = {American Mathematics Competitions},
  year         = {2023},
  note         = {Accessed: 2025-09-17},
  url          = {https://www.maa.org/math-competitions}
}

@misc{zhang2025lessons,
      title={The Lessons of Developing Process Reward Models in Mathematical Reasoning}, 
      author={Zhenru Zhang and Chujie Zheng and Yangzhen Wu and Beichen Zhang and Runji Lin and Bowen Yu and Dayiheng Liu and Jingren Zhou and Junyang Lin},
      year={2025},
      eprint={2501.07301},
      archivePrefix={arXiv},
      primaryClass={cs.CL},
      url={https://arxiv.org/abs/2501.07301}, 
}

@misc{li2025torl,
      title={ToRL: Scaling Tool-Integrated RL}, 
      author={Xuefeng Li and Haoyang Zou and Pengfei Liu},
      year={2025},
      eprint={2503.23383},
      archivePrefix={arXiv},
      primaryClass={cs.CL},
      url={https://arxiv.org/abs/2503.23383}, 
}

@misc{qian2025toolrl,
      title={ToolRL: Reward is All Tool Learning Needs}, 
      author={Cheng Qian and Emre Can Acikgoz and Qi He and Hongru Wang and Xiusi Chen and Dilek Hakkani-Tür and Gokhan Tur and Heng Ji},
      year={2025},
      eprint={2504.13958},
      archivePrefix={arXiv},
      primaryClass={cs.LG},
      url={https://arxiv.org/abs/2504.13958}, 
}

@misc{shao2024deepseekmath,
      title={DeepSeekMath: Pushing the Limits of Mathematical Reasoning in Open Language Models}, 
      author={Zhihong Shao and Peiyi Wang and Qihao Zhu and Runxin Xu and Junxiao Song and Xiao Bi and Haowei Zhang and Mingchuan Zhang and Y. K. Li and Y. Wu and Daya Guo},
      year={2024},
      eprint={2402.03300},
      archivePrefix={arXiv},
      primaryClass={cs.CL},
      url={https://arxiv.org/abs/2402.03300}, 
}

@misc{zhang2025critiquegrpo,
      title={Critique-GRPO: Advancing LLM Reasoning with Natural Language and Numerical Feedback}, 
      author={Xiaoying Zhang and Hao Sun and Yipeng Zhang and Kaituo Feng and Chaochao Lu and Chao Yang and Helen Meng},
      year={2025},
      eprint={2506.03106},
      archivePrefix={arXiv},
      primaryClass={cs.CL},
      url={https://arxiv.org/abs/2506.03106}, 
}

@misc{cui2025prime,
      title={Process Reinforcement through Implicit Rewards}, 
      author={Ganqu Cui and Lifan Yuan and Zefan Wang and Hanbin Wang and Wendi Li and Bingxiang He and Yuchen Fan and Tianyu Yu and Qixin Xu and Weize Chen and Jiarui Yuan and Huayu Chen and Kaiyan Zhang and Xingtai Lv and Shuo Wang and Yuan Yao and Xu Han and Hao Peng and Yu Cheng and Zhiyuan Liu and Maosong Sun and Bowen Zhou and Ning Ding},
      year={2025},
      eprint={2502.01456},
      archivePrefix={arXiv},
      primaryClass={cs.LG},
      url={https://arxiv.org/abs/2502.01456}, 
}

@inproceedings{yuan2025reinforce,
  title={Reinforce LLM Reasoning through Multi-Agent Reflection},
  author={Yurun Yuan and Tengyang Xie},
  booktitle={International Conference on Machine Learning (ICML)},
  year={2025},
  url={https://openreview.net/forum?id=6k3oFS3Lbl}
}

@inproceedings{zhang-xiong-2025-debate4math,
  title     = {{Debate4MATH}: Multi-Agent Debate for Fine-Grained Reasoning in Math},
  author    = {Zhang, Shaowei and Xiong, Deyi},
  booktitle = {Proceedings of the 63rd Annual Meeting of the Association for Computational Linguistics (Volume 5: Findings of ACL)},
  month     = {July},
  year      = {2025},
  address   = {Vancouver, Canada},
  publisher = {Association for Computational Linguistics},
  url       = {https://aclanthology.org/2025.findings-acl.862},
  pages     = {14524--14537}
}

@article{zhou2025reso,
  title={Reso: A reward-driven self-organizing llm-based multi-agent system for reasoning tasks},
  author={Zhou, Hongyi and Geng, Haoran and Xue, Xingcheng and Kang, Lei and Qin, Yuhang and Wang, Zihan and others},
  journal={arXiv preprint arXiv:2503.02390},
  year={2025},
  url={https://arxiv.org/abs/2503.02390}
}

@inproceedings{motwani2025malt,
  title={MALT: Improving Reasoning with Multi-Agent {LLM} Training},
  author={Motwani, Sumeet Ramesh and Smith, Chandler and Das, Rocktim Jyoti and Rafailov, Rafael and Laptev, Ivan and Torr, Philip and Pizzati, Fabio and Clark, Ronald and de Witt, Christian Schroeder},
  booktitle={International Conference on Learning Representations (ICLR)},
  year={2025},
  url={https://openreview.net/forum?id=jXP9bgFack}
}

@article{wang2023mathcoder,
  title={MathCoder: Seamless Code Integration in {LLMs} for Enhanced Mathematical Reasoning},
  author={Wang, Ke and Ren, Houxing and Zhou, Aojun and Lu, Zimu and Luo, Sichun and Shi, Weikang and Zhang, Renrui and Song, Linqi and Zhan, Mingjie and Li, Hongsheng},
  journal={arXiv preprint arXiv:2310.03731},
  year={2023},
  url={https://arxiv.org/abs/2310.03731}
}

@misc{jin2025searchr1,
      title={Search-R1: Training LLMs to Reason and Leverage Search Engines with Reinforcement Learning}, 
      author={Bowen Jin and Hansi Zeng and Zhenrui Yue and Jinsung Yoon and Sercan Arik and Dong Wang and Hamed Zamani and Jiawei Han},
      year={2025},
      eprint={2503.09516},
      archivePrefix={arXiv},
      primaryClass={cs.CL},
      url={https://arxiv.org/abs/2503.09516}, 
}

@inproceedings{wu2025agentic,
  title     = {Agentic Reasoning: A Streamlined Framework for Enhancing {LLM} Reasoning with Agentic Tools},
  author    = {Wu, Junde and Zhu, Jiayuan and Liu, Yuyuan and Xu, Min and Jin, Yueming},
  booktitle = {Proceedings of the 63rd Annual Meeting of the Association for Computational Linguistics},
  year      = {2025},
  month     = {July},
  address   = {Vancouver, Canada},
  publisher = {Association for Computational Linguistics},
  url       = {https://aclanthology.org/2025.acl-long.1383}
}

@article{gou2023tora,
  title={ToRA: A Tool-Integrated Reasoning Agent for Mathematical Problem Solving},
  author={Gou, Zhibin and Shao, Zhihong and Gong, Yeyun and Shen, Yelong and Yang, Yujiu and Huang, Minlie and Duan, Nan and Chen, Weizhu},
  journal={arXiv preprint arXiv:2309.17452},
  year={2023},
  url={https://arxiv.org/abs/2309.17452}
}

@article{openai2024o1,
  title={OpenAI o1 System Card},
  author={Jaech, Aaron and Kalai, Adam and Lerer, Adam and Richardson, Adam and El-Kishky, Ahmed and Low, Aiden and Helyar, Alec and Madry, Aleksander and Carney, Alex and Beutel, Alex and others},
  journal={arXiv preprint arXiv:2412.16720},
  year={2024},
  url={https://arxiv.org/abs/2412.16720}
}

@article{han2024llmmas,
  title={{LLM} Multi-Agent Systems: Challenges and Open Problems},
  author={Han, Shanshan and Zhang, Qifan and Yao, Yuhang and Jin, Weizhao and Xu, Zhaozhuo and He, Chaoyang},
  journal={arXiv preprint arXiv:2402.03578},
  year={2024},
  url={https://arxiv.org/abs/2402.03578}
}

@article{tran2025multi,
  title={Multi-Agent Collaboration Mechanisms: A Survey of {LLMs}},
  author={Tran, Khanh-Tung and Dao, Dung and Nguyen, Minh-Duong and Pham, Quoc-Viet and O'Sullivan, Barry and Nguyen, Hoang D.},
  journal={arXiv preprint arXiv:2501.06322},
  year={2025},
  url={https://arxiv.org/abs/2501.06322}
}

@article{lei2024macm,
  title={{MACM}: Utilizing a Multi-Agent System for Condition Mining in Solving Complex Mathematical Problems},
  author={Lei, Bin and Zhang, Yi and Zuo, Shan and Payani, Ali and Ding, Caiwen},
  journal={arXiv preprint arXiv:2404.04735},
  year={2024},
  url={https://arxiv.org/abs/2404.04735}
}

@article{wang2024mixture,
  title={Mixture-of-Agents Enhances Large Language Model Capabilities},
  author={Wang, Junlin and Wang, Jue and Athiwaratkun, Ben and Zhang, Ce and Zou, James},
  journal={arXiv preprint arXiv:2406.04692},
  year={2024},
  url={https://arxiv.org/abs/2406.04692}
}

@article{zhuge2024language,
  title={Language Agents as Optimizable Graphs},
  author={Zhuge, Mingchen and Wang, Wenyi and Kirsch, Louis and Faccio, Francesco and Khizbullin, Dmitrii and Schmidhuber, J{\"u}rgen},
  journal={arXiv preprint arXiv:2402.16823},
  year={2024},
  url={https://arxiv.org/abs/2402.16823}
}

@article{zhang2025gdesigner,
  title={{G-Designer}: Architecting Multi-Agent Communication Topologies via Graph Neural Networks},
  author={Zhang, Guibin and Yue, Yanwei and Sun, Xiangguo and Wan, Guancheng and Yu, Miao and Fang, Junfeng and Wang, Kun and Chen, Tianlong and Cheng, Dawei},
  journal={arXiv preprint arXiv:2410.11782},
  year={2025},
  url={https://arxiv.org/abs/2410.11782}
}

@article{ouyang2022training,
  title={Training language models to follow instructions with human feedback},
  author={Ouyang, Long and Wu, Jeffrey and Jiang, Xu and Almeida, Diogo and Wainwright, Carroll and Mishkin, Paul and others},
  journal={Advances in Neural Information Processing Systems},
  volume={35},
  pages={27730--27744},
  year={2022}
}

@misc{lightman2023letsverifystepstep,
      title={Let's Verify Step by Step}, 
      author={Hunter Lightman and Vineet Kosaraju and Yura Burda and Harri Edwards and Bowen Baker and Teddy Lee and Jan Leike and John Schulman and Ilya Sutskever and Karl Cobbe},
      year={2023},
      eprint={2305.20050},
      archivePrefix={arXiv},
      primaryClass={cs.LG},
      url={https://arxiv.org/abs/2305.20050}, 
}

@inproceedings{luo2024wizardmath,
  title={{WizardMath}: Empowering Mathematical Reasoning for Large Language Models via Reinforced Evol-Instruct},
  author={Luo, Haipeng and Sun, Qingfeng and Xu, Can and Zhao, Pu and Lou, Jian-Guang and Tao, Chongyang and Geng, Xiubo and Lin, Qingwei and Chen, Shifeng and Tang, Yansong and Zhang, Dongmei},
  booktitle={The Twelfth International Conference on Learning Representations (ICLR)},
  year={2024},
  url={https://openreview.net/forum?id=mMPMHWOdOy}
}

@article{schulman2017proximal,
  title={Proximal policy optimization algorithms},
  author={Schulman, John and Wolski, Filip and Dhariwal, Prafulla and Radford, Alec and Klimov, Oleg},
  journal={arXiv preprint arXiv:1707.06347},
  year={2017}
}

@article{rafailov2023direct,
  title={Direct preference optimization: Your language model is secretly a reward model},
  author={Rafailov, Rafael and Sharma, Archit and Mitchell, Eric and Manning, Christopher D and Ermon, Stefano and Finn, Chelsea},
  journal={Advances in Neural Information Processing Systems},
  volume={36},
  pages={53728--53741},
  year={2023}
}

@article{yu2024steptool,
  title={StepTool: A step-grained reinforcement learning framework for tool learning in LLMs},
  author={Yu, Yuanqing and Wang, Zhefan and Ma, Weizhi and Guo, Zhicheng and Zhan, Jingtao and Wang, Shuai and Wu, Chuhan and Guo, Zhiqiang and Zhang, Min},
  journal={arXiv preprint arXiv:2410.07745},
  year={2024}
}
\bibliographystyle{iclr2025_conference}

\appendix
\clearpage
\section{motivation prompts}
\label{app:motivation_prompt}
\paragraph{Reasoning-Only Prompt.}
The following prompt is used during all reasoning-only inference experiments:

\begin{tcolorbox}[
  colframe=blue!75!black, 
  colback=blue!5, 
  title=Reasoning-Only Prompt, 
  coltitle=white, 
  fonttitle=\bfseries 
]
\textbf{System:} You are an IMO medalist mathematician. 

\textbf{User:}

Question:
\texttt{<Question Content>} 

Instructions:

1. Think step-by-step in natural language to derive a complete solution.

2. Provide ONLY the final mathematical answer in the last line, formatted as:

ANSWER: \texttt{<Your Answer>}

3. Do NOT write or reference any code.  
Stop after giving the answer.
\end{tcolorbox}

\paragraph{Reasoning with Code Prompt.}
The following prompt is used during all reasoning with code inference experiments:

\begin{tcolorbox}[
  colframe=blue!75!black,
  colback=blue!5,
  title=Reasoning with Code Prompt,
  coltitle=white,
  fonttitle=\bfseries
]
\textbf{System:}  You are an IMO medalist mathematician who can also run Python (with sympy, itertools, math, random).

\textbf{User:}

Question:
\texttt{<Question Content>}

Instructions:

1. Think step-by-step in natural language.

2. Whenever useful, place Python code inside ```python ...``` cells. The code will be executed; you can refer to its output.

3. After you finish reasoning, output your final answer on a separate line:
ANSWER: \texttt{<Your Answer>} 
\end{tcolorbox}

\paragraph{Prompt for Reasoning Path Critique.}
We adopt a prompt to enable DeepSeek-R1 to evaluate the correctness of generated responses using the above two prompting regimes:

\begin{tcolorbox}[
  colframe=blue!75!black,
  colback=blue!5,
  title=Prompt for Reasoning Path Critique,
  coltitle=white,
  fonttitle=\bfseries
]
\textbf{System:} You are an experienced automatic grading assistant. Only output the result for the question in the specified format.

\textbf{User:}

I have a question: \texttt{<Question Content>}

Below are 5 different answers to this question, all from the same small model.

For each answer, please judge whether the answer contains a correct approach (i.e., Is the line of thinking or analysis correct? It doesn't need to be a fully correct final answer, just whether the reasoning process is on the right track).

Count how many out of the 5 responses have a correct approach, and ONLY output the number of correct reasoning answers in the format: \#number\# (for example, \#3\# means 3 out of 5 answers are correct in reasoning).

Do not provide any explanations, just output the result in the requested format.

Here are the answers:

\texttt{<Answer List>}
\end{tcolorbox}

\clearpage
\section{Motivation qualitative example}
\label{app:motivation_example}
This example demonstrates a comparison between two different prompting methods used to solve a medium-hard (level 4) math problem with the Qwen2.5-Math-1.5B-Instruct model: 

\begin{tcolorbox}[
  enhanced,
  colback=yellow!10,
  colframe=yellow!50!black,
  arc=3mm,
  boxrule=1pt,
  rounded corners,
  title={\normalsize\textbf{Question}},  
  fonttitle=\normalsize\bfseries,          
  coltitle=black,
  attach title to upper=\quad,
  left=4pt,         
  right=4pt,        
  top=4pt,          
  bottom=4pt,       
  toptitle=2pt,     
  bottomtitle=2pt   
] 

Given
\[
\mathbf{A} = \begin{pmatrix} 3 & 7 \\ -1 & -2 \end{pmatrix},
\]
compute $\mathbf{A}^{27} + \mathbf{A}^{31} + \mathbf{A}^{40}$.
\end{tcolorbox}

\begin{tcolorbox}[
  enhanced,
  colback=yellow!10,
  colframe=yellow!50!black,
  arc=3mm,
  boxrule=1pt,
  rounded corners,
  title={\normalsize\textbf{Solution (Ground Truth)}},  
  fonttitle=\normalsize\bfseries,          
  coltitle=black,
  attach title to upper=\quad,
  left=4pt,         
  right=4pt,        
  top=4pt,          
  bottom=4pt,       
  toptitle=2pt,     
  bottomtitle=2pt   
] 

Note that
\[
\mathbf{A}^2 = \begin{pmatrix} 2 & 7 \\ -1 & -3 \end{pmatrix}
\]
and
\[
\mathbf{A}^3 = \mathbf{A} \mathbf{A}^2 = \begin{pmatrix} -1 & 0 \\ 0 & -1 \end{pmatrix} = -\mathbf{I}.
\]
Then
\begin{align*}
\mathbf{A}^{27} + \mathbf{A}^{31} + \mathbf{A}^{40} &= (\mathbf{A}^3)^9 + (\mathbf{A}^3)^{10} \mathbf{A} + (\mathbf{A}^3)^{13} \mathbf{A} \\
&= (-\mathbf{I})^9 + (-\mathbf{I})^{10} \mathbf{A} + (-\mathbf{I})^{13} \mathbf{A} \\
&= -\mathbf{I} + \mathbf{A} - \mathbf{A} \\
&= -\mathbf{I} = \boxed{\begin{pmatrix} -1 & 0 \\ 0 & -1 \end{pmatrix}}.
\end{align*}
\end{tcolorbox}

\begin{tcolorbox}[
  enhanced,
  colback=purple!10,
  colframe=purple!50!black,
  arc=3mm,
  boxrule=1pt,
  rounded corners,
  title={\normalsize\textbf{Solution 1: Incorrect Response (Reasoning with Code)}},  
  fonttitle=\normalsize\bfseries,          
  coltitle=black,
  attach title to upper=\quad,
  left=4pt,         
  right=4pt,        
  top=4pt,          
  bottom=4pt,       
  toptitle=2pt,     
  bottomtitle=2pt   
]

To solve the problem, we need to compute $\mathbf{A}^{27} + \mathbf{A}^{31} + \mathbf{A}^{40}$ where $\mathbf{A} = \begin{pmatrix} 3 & 7 \\ -1 & -2 \end{pmatrix}$.

The steps are as follows:

\begin{enumerate}
    \item Find the eigenvalues and eigenvectors of $\mathbf{A}$.
    \item Diagonalize the matrix $\mathbf{A}$ if possible, i.e., express $\mathbf{A}$ as $\mathbf{PDP}^{-1}$, where $\mathbf{D}$ is a diagonal matrix of eigenvalues and $\mathbf{P}$ is a matrix of corresponding eigenvectors.
    \item Compute $\mathbf{A}^{27}$, $\mathbf{A}^{31}$, and $\mathbf{A}^{40}$ using the diagonalized form.
    \item Sum the resulting matrices.
\end{enumerate}

Let's implement this in Python using sympy.

\texttt{```python}

\begin{lstlisting}[language=Python, caption={}]
import sympy as sp

# Define the matrix A
A = sp.Matrix([[3, 7], [-1, -2]])

# Find the eigenvalues and eigenvectors
# Note: diagonalize() returns (P, D) where A = P * D * P.inv()
P, D = A.diagonalize()

# Compute A^27, A^31, and A^40 using the diagonalized form
A_27 = P * D**27 * P.inv()
A_31 = P * D**31 * P.inv()
A_40 = P * D**40 * P.inv()

# Compute the final result
result = A_27 + A_31 + A_40

# Print the final result
print(result)
\end{lstlisting}
\texttt{```}

\texttt{```output}

\begin{lstlisting}[style=outputstyle, caption={}]
sympy.matrices.common.MatrixError: Matrix is not diagonalizable
\end{lstlisting}

\texttt{```}

The error indicates that the matrix $\mathbf{A}$ is not diagonalizable. This means that we cannot directly diagonalize $\mathbf{A}$ and use the diagonalization method to compute $\mathbf{A}^{27}$, $\mathbf{A}^{31}$, and $\mathbf{A}^{40}$.

Instead, we can use the Cayley-Hamilton theorem, which states that a matrix satisfies its own characteristic equation. The characteristic polynomial of $\mathbf{A}$ is given by:
\[
\det(\mathbf{A} - \lambda \mathbf{I}) = \begin{vmatrix} 3 - \lambda & 7 \\ -1 & -2 - \lambda \end{vmatrix} = (3 - \lambda)(-2 - \lambda) + 7 = \lambda^2 - \lambda + 2.
\]

According to the Cayley-Hamilton theorem, $\mathbf{A}^2 - \mathbf{A} + \mathbf{I} = \mathbf{0}$, which implies:
\[
\mathbf{A}^2 = \mathbf{A} - \mathbf{I}.
\]
Using this relation, we can express higher powers of $\mathbf{A}$ in terms of $\mathbf{A}$ and $\mathbf{I}$.

Let's use this relation to compute $\mathbf{A}^{27}$, $\mathbf{A}^{31}$, and $\mathbf{A}^{40}$.

\begin{enumerate}
    \item Compute $\mathbf{A}^{27}$:
    \[
    \mathbf{A}^{27} = \mathbf{A}^{27 \mod 2} = \mathbf{A}^1 = \mathbf{A}.
    \]
    \item Compute $\mathbf{A}^{31}$:
    \[
    \mathbf{A}^{31} = \mathbf{A}^{31 \mod 2} = \mathbf{A}^1 = \mathbf{A}.
    \]
    \item Compute $\mathbf{A}^{40}$:
    \[
    \mathbf{A}^{40} = \mathbf{A}^{40 \mod 2} = \mathbf{A}^0 = \mathbf{I}.
    \]
\end{enumerate}
Thus, $\mathbf{A}^{27} + \mathbf{A}^{31} + \mathbf{A}^{40} = \mathbf{A} + \mathbf{A}$ \text{...(truncated)}

\end{tcolorbox}

\begin{tcolorbox}[
  enhanced,
  colback=green!10,
  colframe=green!50!black,
  arc=3mm,
  boxrule=1pt,
  rounded corners,
  title={\normalsize\textbf{Solution 2: Correct Response (Reasoning Only)}},  
  fonttitle=\normalsize\bfseries,          
  coltitle=black,
  attach title to upper=\quad,
  left=4pt,         
  right=4pt,        
  top=4pt,          
  bottom=4pt,       
  toptitle=2pt,     
  bottomtitle=2pt   
]

To solve the problem of computing $\mathbf{A}^{27} + \mathbf{A}^{31} + \mathbf{A}^{40}$, we need to first understand the behavior of the matrix $\mathbf{A}$ under exponentiation. We will use the characteristic polynomial of the matrix $\mathbf{A}$ to find its eigenvalues and then use these eigenvalues to find a pattern in the powers of $\mathbf{A}$.

The characteristic polynomial of $\mathbf{A}$ is given by:
\[
\det(\mathbf{A} - \lambda \mathbf{I}) = \begin{vmatrix} 3 - \lambda & 7 \\ -1 & -2 - \lambda \end{vmatrix} = (3 - \lambda)(-2 - \lambda) + 7 = \lambda^2 - \lambda + 1.
\]

The eigenvalues $\lambda_1$ and $\lambda_2$ are the roots of the characteristic polynomial $\lambda^2 - \lambda + 1 = 0$. Solving this quadratic equation, we get:
\[
\lambda = \frac{1 \pm \sqrt{1 - 4}}{2} = \frac{1 \pm \sqrt{-3}}{2} = \frac{1 \pm i\sqrt{3}}{2}.
\]
Let $\lambda_1 = \frac{1 + i\sqrt{3}}{2}$ and $\lambda_2 = \frac{1 - i\sqrt{3}}{2}$.

Since $\mathbf{A}$ is a 2x2 matrix with eigenvalues $\lambda_1$ and $\lambda_2$, we can express $\mathbf{A}$ in the form $\mathbf{A} = \mathbf{P} \mathbf{D} \mathbf{P}^{-1}$, where $\mathbf{D}$ is the diagonal matrix of eigenvalues and $\mathbf{P}$ is the matrix of eigenvectors.

However, instead of explicitly finding $\mathbf{P}$ and $\mathbf{P}^{-1}$, we can use the fact that $\mathbf{A}^n = \mathbf{P} \mathbf{D}^n \mathbf{P}^{-1}$ for any integer $n$.

Since $\mathbf{D}$ is diagonal, $\mathbf{D}^n$ is simply the diagonal matrix with $\lambda_1^n$ and $\lambda_2^n$ on the diagonal.

Observe that $\lambda_1 = e^{i\pi/3}$ and $\lambda_2 = e^{-i\pi/3}$ are primitive 6th roots of unity. Therefore:
$$
\lambda_1^3 = e^{i\pi} = -1, \quad \lambda_2^3 = e^{-i\pi} = -1.
$$

Since $\mathbf{A}^3$ has eigenvalues $\lambda_1^3 = -1$, $\lambda_2^3 = -1$, and since $\mathbf{A}^3$ is diagonalizable, it must be:
$$
\mathbf{A}^3 = -\mathbf{I}.
$$

More precisely:
$$
\mathbf{A}^n =
\begin{cases}
(-1)^k \mathbf{I}, & n = 3k, \\
(-1)^k \mathbf{A}, & n = 3k+1, \\
(-1)^k \mathbf{A}^2, & n = 3k+2.
\end{cases}
$$
\text{(truncated)}
\end{tcolorbox}

\textbf{Analysis.} The initial implementation of Solution 1 follows a mathematically sound strategy: attempt to diagonalize the matrix $\mathbf{A}$ and use the decomposition $\mathbf{A} = \mathbf{PDP}^{-1}$ to compute high powers efficiently via $\mathbf{A}^n = \mathbf{P D}^n \mathbf{P}^{-1}$. The Python code using \texttt{sympy} is syntactically correct and logically coherent. This approach would succeed if $\mathbf{A}$ were diagonalizable over the real numbers. However, the code raises the exception: "sympy.matrices.common.MatrixError: Matrix is not diagonalizable"
This error disrupts the solution path and triggers a cascade of flawed reasoning. Although the matrix 
$$
\mathbf{A} = \begin{pmatrix} 3 & 7 \\ -1 & -2 \end{pmatrix}
$$
has distinct eigenvalues $\lambda = \frac{1 \pm i\sqrt{3}}{2}$ (roots of $\lambda^2 - \lambda + 1 = 0$), it is diagonalizable \textit{over the complex numbers}. The error arises because \texttt{diagonalize()} may fail to find a real eigenvector matrix, not because diagonalization is impossible in principle.
Instead of handling this gracefully—such as by manually computing eigenvectors over $\mathbb{C}$ or switching to an alternative method—the solution incorrectly interprets the error as a complete dead end. This leads to a series of hallucinated steps:

\begin{itemize}
    \item The characteristic polynomial is incorrectly stated as $\lambda^2 - \lambda + 2$, though the subsequent derivation $\mathbf{A}^2 = \mathbf{A} - \mathbf{I}$ implies the correct polynomial $\lambda^2 - \lambda + 1$ was intended.
    \item Most critically, the solution \textbf{fabricates a false periodicity}, claiming:
    $$
    \mathbf{A}^{27} = \mathbf{A}^{27 \mod 2} = \mathbf{A}, \quad \mathbf{A}^{40} = \mathbf{I}, \quad \text{etc.}
    $$
    This ``mod 2 reduction'' has no mathematical basis and is a hallucination induced by the earlier computational failure.
\end{itemize}

In contrast, Solution 2 presents a sophisticated and mathematically sound approach to computing the problem, even without code invocation and execution. It avoids brute-force computation by leveraging the algebraic structure of the matrix through eigenvalues and periodicity. The solution is truncated (due to the limit on generation max tokens), but the framework is complete.

In summary, while the initial code logic is correct, the Solution 1 lacks \textbf{mathematical robustness} in handling the exception. The interruption caused by the diagonalization error derails the reasoning, resulting in hallucinated algebraic rules and a fundamentally incorrect final answer.

\clearpage
\section{Successful Rollout Example}
\label{app:method_example}
\begin{tcolorbox}[
  enhanced,
  colback=yellow!10,
  colframe=yellow!50!black,
  arc=3mm,
  boxrule=1pt,
  rounded corners,
  title={\normalsize\textbf{Question}},  
  fonttitle=\normalsize\bfseries,          
  coltitle=black,
  attach title to upper=\quad,
  left=4pt,         
  right=4pt,        
  top=4pt,          
  bottom=4pt,       
  toptitle=2pt,     
  bottomtitle=2pt   
] 

If $f(x) = \frac{3x-2}{x-2}$, what is the value of $f(-2) +f(-1)+f(0)$? Express your answer as a common fraction.
\end{tcolorbox}

\begin{tcolorbox}[
  enhanced,
  colback=yellow!10,
  colframe=yellow!50!black,
  arc=3mm,
  boxrule=1pt,
  rounded corners,
  title={\normalsize\textbf{Solution}},  
  fonttitle=\normalsize\bfseries,          
  coltitle=black,
  attach title to upper=\quad,
  left=4pt,         
  right=4pt,        
  top=4pt,          
  bottom=4pt,       
  toptitle=2pt,     
  bottomtitle=2pt   
] 

$f(-2)+f(-1)+f(0)=\frac{3(-2)-2}{-2-2}+\frac{3(-1)-2}{-1-2}+\frac{3(0)-2}{0-2}=\frac{-8}{-4}+\frac{-5}{-3}+\frac{-2}{-2}=2+\frac{5}{3}+1=\boxed{\frac{14}{3}}$
\end{tcolorbox}

\begin{tcolorbox}[
  enhanced,
  colback=blue!10,
  colframe=blue!50!black,
  arc=3mm,
  boxrule=1pt,
  rounded corners,
  title={\normalsize\textbf{Rollout Example}},  
  fonttitle=\normalsize\bfseries,          
  coltitle=black,
  attach title to upper=\quad,
  left=4pt,         
  right=4pt,        
  top=4pt,          
  bottom=4pt,       
  toptitle=2pt,     
  bottomtitle=2pt   
]

Please integrate natural language reasoning with programs to solve the problem above, and put your final answer within boxed{} Problem: If $f(x) = \frac{3x-2}{x-2}$, what is the value of $f(-2) +f(-1)+f(0)$? Express your answer as a common fraction. To solve the problem, we need to evaluate the function \( f(x) = \frac{3x-2}{x-2} \) at \( x = -2 \), \( x = -1 \), and \( x = 0 \), and then sum the results. Let's break it down step-by-step: 1. Evaluate \( f(-2) \): \[ f(-2) = \frac{3(-2) - 2}{-2 - 2} = \frac{-6 - 2}{-4} = \frac{-8}{-4} = 2 \] 2. Evaluate \( f(-1) \): \[ f(-1) = \frac{3(-1) - 2}{-1 - 2} = \frac{-3 - 2}{-3} = \frac{-5}{-3} = \frac{5}{3} \] 3. Evaluate \( f(0) \): \[ f(0) = \frac{3(0) - 2}{0 - 2} = \frac{-2}{-2} = 1 \] 4. Sum the results: \[ f(-2) + f(-1) + f(0) = 2 + \frac{5}{3} + 1 = 3 + \frac{5}{3} = \frac{9}{3} + \frac{5}{3} = \frac{14}{3} \] Now, let's confirm this with Python code. ```python This Python script defines a function `f` that calculates the value of \((3x - 2)/(x - 2)\). It then evaluates this function at \(x = -2\), \(x = -1\), and \(x = 0\) to find their respective values. The sum of these three values is printed as the final result. ``` ```output 4.666666666666667 ``` The output from the code is approximately \(4.6667\), which corresponds to the fraction \(\frac{14}{3}\). Thus, the value of \( f(-2) + f(-1) + f(0) \) is \(\boxed{\frac{14}{3}}\).
\end{tcolorbox}

\clearpage
\section{MSARL Dual-Agent Optimization}
\label{app:refinement_process}
\begin{minipage}{1\textwidth}
\label{alg:refinement_process}
\begin{algorithm}[H]
\caption{\texttt{MSARL}: Dual-Agent Optimization via Grouped Rollouts}
\begin{algorithmic}[1]
\State \textbf{Input:} the reasoning agent $\pi_{\text{reason}}$, the tool agent $\pi_{\text{tool}}$, the given question set $Q$,  the ground truth $GT$, a reward system $\mathcal{R}$
\State \textbf{Initialize:} reasoning policy parameters $\theta_{\text{reason}}$, tool policy parameters $\theta_{\text{tool}}$.
\State \textbf{Hyperparameters:}  learning rates of reasoning and tool policies $\alpha_{\text{reason}}, \alpha_{\text{tool}}$; number of reasoning trajectories per question $m$, number of tool interpretations per trajectory $n$, maximum tool calls $C$, PPO clip value $\epsilon$.
\For{each training iteration}
    \State Sample $\{q_j\} \subset Q$.
    \State the reasoning and tool datasets: \\$D_{\text{reason}} \leftarrow \emptyset$, $D_{\text{tool}} \leftarrow \emptyset$.
\Statex \textcolor{gray}{\textit{// --- Phase 1: Data Generation ---}}
    \For{each $q \in \{q_j\}$}
        \State Sample $\{\tau_k\}_{k=1}^m \sim \pi_{\text{reason}}(\cdot|q)$.
        \For{each $\tau_k$}
            \State Let $(t, o)_{\text{last}} \in \tau_k$.
            \State Sample $\{i_k^{(j)}\}_{j=1}^n \sim \pi_{\text{tool}}(\cdot | (t, o)_{\text{last}})$.
            \State For each $i_k^{(j)}$, \\generate $O_k^{(j)} \leftarrow \pi_{\text{reason}}(\tau_k, i_k^{(j)})$.
            \State Store $(\tau_k, \{ (i_k^{(j)}, O_k^{(j)}) \}_{j=1}^n)$.
        \EndFor
    \EndFor
\Statex \textcolor{gray}{\textit{// --- Phase 2: Advantage Estimation ---}}
    \For{each $(\tau_k, \{ (i_k^{(j)}, O_k^{(j)}) \}_{j=1}^n)$}
        \State $R_k^{(j)} \leftarrow \mathcal{R}(O_k^{(j)}, GT)$.
        \State $\mu_k \leftarrow \frac{1}{n}\sum_j R_k^{(j)}$; $\sigma_k \leftarrow \sqrt{\frac{1}{n}\sum_j (R_k^{(j)}-\mu_k)^2}$.
        \For{$j=1$ to $n$}
            \State $A_k^{(j)} \leftarrow (R_k^{(j)} - \mu_k) / (\sigma_k + \eta)$.
            \State $D_{\text{tool}} \leftarrow D_{\text{tool}} \cup \{ (i_k^{(j)}, A_k^{(j)}) \}$.
        \EndFor
        \State $\bar{A}_k \leftarrow \frac{1}{n}\sum_j A_k^{(j)}$.
        \State $D_{\text{reason}} \leftarrow D_{\text{reason}} \cup \{ (\tau_k, \bar{A}_k) \}$.
    \EndFor
\Statex \textcolor{gray}{\textit{// ---  Phase 3: Policy Update ---}}
    \State $\theta_{\text{tool}} \leftarrow \theta_{\text{tool}} + \alpha_{\text{tool}} \nabla_{\theta_{\text{tool}}} \mathcal{J}(\theta_{\text{tool}} | D_{\text{tool}})$.
    \State $\theta_{\text{reason}} \leftarrow \theta_{\text{reason}} + \alpha_{\text{reason}} \nabla_{\theta_{\text{reason}}} \mathcal{J}(\theta_{\text{reason}} | D_{\text{reason}})$.
\EndFor
\end{algorithmic}
\end{algorithm}
\end{minipage}

\clearpage
\section{Statement on the Use of Large Language Models (LLMs)}
\label{app:llm_use}
In the preparation of this paper, we utilized a Large Language Model (LLM) as a writing assistant to help with language polishing, grammar checking, and text optimization. The use of the LLM was strictly limited to improving the paper's readability and clarity of expression, and was not used to generate any research content, core arguments, or data.

We hereby declare that all research originality, core ideas, experimental methods, results, and conclusions in this paper were developed and finalized by the authors independently. The authors take full responsibility for all aspects of the submission and guarantee its truthfulness and accuracy.

\end{document}